\documentclass[numbers,square]{article}
\usepackage{booktabs}

\usepackage{amsmath}
\usepackage{wrapfig}

\usepackage[preprint]{corl_2025} 
\usepackage{amsmath,amssymb,amsfonts}

\usepackage{hyperref}       
\hypersetup{
    citecolor=orange,    
}
\usepackage{url}            
\usepackage{algorithmic}
\usepackage{graphicx}
\usepackage{textcomp}
\usepackage[table,x11names]{xcolor}
\usepackage{amsmath}
\usepackage[ruled,vlined]{algorithm2e}
\usepackage{hyperref}
\usepackage{xurl}
\usepackage{placeins}
\usepackage[subtle]{savetrees} %
\usepackage{cleveref}

\title{RoboMonkey: Scaling Test-Time Sampling~and Verification for Vision-Language-Action Models}

\author{
\begin{tabular}{ccccc}
Jacky Kwok\textsuperscript{1} &
Christopher Agia\textsuperscript{1,†} &
Rohan Sinha\textsuperscript{1,†} &
Matt Foutter\textsuperscript{1,†} \\
Shulu Li\textsuperscript{2} &
Ion Stoica\textsuperscript{2} &
Azalia Mirhoseini\textsuperscript{1} &
Marco Pavone\textsuperscript{1,3} &
\end{tabular} \\[0.45cm]
\textcolor{black}{\textsuperscript{1}Stanford University} \quad
\textcolor{black}{\textsuperscript{2}UC Berkeley} \quad
\textcolor{black}{\textsuperscript{3}NVIDIA Research}
}

\begin{document}
\maketitle
\begingroup
\renewcommand\thefootnote{\textsuperscript{†}}
\footnotetext{Equal contribution. Correspondence to: jackykwok@stanford.edu}
\endgroup

\vspace{-0.35in}
\begin{center}
\href{https://robomonkey-vla.github.io}{\textcolor{orange}{https://robomonkey-vla.github.io}}
\end{center}
\vspace{-0.06in}
\begin{abstract}
Vision-Language-Action (VLA) models have demonstrated remarkable capabilities in visuomotor control, yet ensuring their robustness in unstructured real-world environments remains a persistent challenge. In this paper, we investigate test-time scaling through the lens of sampling and verification as means to enhance the robustness and generalization of VLAs. We first demonstrate that the relationship between action error and the number of generated samples follows an exponentiated power law across a range of VLAs, indicating the existence of inference-time scaling laws. Building on these insights, we introduce RoboMonkey, a test-time scaling framework for VLAs. At deployment, RoboMonkey samples a small set of actions from a VLA, applies Gaussian perturbation and majority voting to construct an action proposal distribution, and then uses a Vision Language Model (VLM)-based verifier to select the optimal action. We propose a synthetic data generation pipeline for training such VLM-based action verifiers, and demonstrate that scaling the synthetic dataset consistently improves verification and downstream accuracy. Through extensive simulated and hardware experiments, we show that pairing existing VLAs with RoboMonkey yields significant performance gains, achieving a 25\% absolute improvement on out-of-distribution tasks and 9\% on in-distribution tasks. Additionally, when adapting to new robot setups, we show that fine-tuning both VLAs and action verifiers yields a 7\% performance increase compared to fine-tuning VLAs alone.

\end{abstract}


\vspace{-0.1in}
\section{Introduction}

Foundation models, pre-trained on extensive internet-scale data, have demonstrated significant potential in robotics domains. Recent advancements in Vision-Language-Action (VLA) models~\cite{black2024pi0visionlanguageactionflowmodel, kim2024openvla, driess2023palmeembodiedmultimodallanguage, brohan2022rt, brohan2023rt} have shown that scaling up training compute on large-scale robotics datasets~\cite{khazatsky2024droid,o2023open} can improve their capabilities and generalization. Despite these advancements, VLAs exhibit diverse failure modes during deployment~\cite{agia2024unpackingfailuremodesgenerative, sinha2024realtimeanomalydetectionreactive}, such as imprecise grasping, task progression failure, and collision with surrounding objects. Addressing these limitations could accelerate the deployment of robots in unstructured real-world environments. 

Efforts to improve the robustness and generalization of VLAs have gradually shifted from the pre-training to the post-training phase. In the pre-training stage, previous work emphasizes scaling up data collection~\cite{zhou2024autonomousimprovementinstructionfollowing, o2023open, walke2023bridgedata}, optimizing training data mixtures~\cite{hejna2024remixoptimizingdatamixtures, kim2024openvla}, and developing model architectures~\cite{chi2023diffusion, black2024pi0visionlanguageactionflowmodel, cheang2024gr2generativevideolanguageactionmodel,zhao2023learningfinegrainedbimanualmanipulation} that can be effectively adapted for robot control. More recently, we have observed a paradigm shift toward developments in the post-training phase, e.g., fine-tuning VLAs for multi-step reasoning with chain-of-thought~\cite{zawalski2025roboticcontrolembodiedchainofthought, clark2025actionfreereasoningpolicygeneralization, zhao2025cotvlavisualchainofthoughtreasoning} and aligning VLAs with preferences~\cite{zhang2025grapegeneralizingrobotpolicy, zhang2025safevlasafetyalignmentvisionlanguageaction, li2025finetuninggenerativetrajectorymodel}. However, beyond pre-training and post-training, less attention has been paid to scaling the amount of compute used during deployment, as VLA models are typically designed to generate a single action chunk per observation.

Humans naturally allocate more time when encountering challenging problems. For Large Language Models (LLMs), this principle has been validated by applying additional compute at test time~\cite{brown2024large,snell2024scaling, saad2024archon,chen2024more, song2024good, li2025stesttimescaling}. Specifically, repeatedly sampling candidate solutions from a model has been shown to enhance the capabilities of LLMs across multiple domains, including mathematics, coding, chat, and summarization~\cite{brown2024large,chen2024alphamath, cobbe2021training}. This raises the question of whether test-time scaling with repeated sampling may also benefit robotics. More precisely, we ask in this work: given an observation and task instruction, can we improve the precision and robustness of VLAs by repeatedly sampling and verifying actions at deployment?

\begin{figure*}
    \centering
    \includegraphics[width=1.0\textwidth]{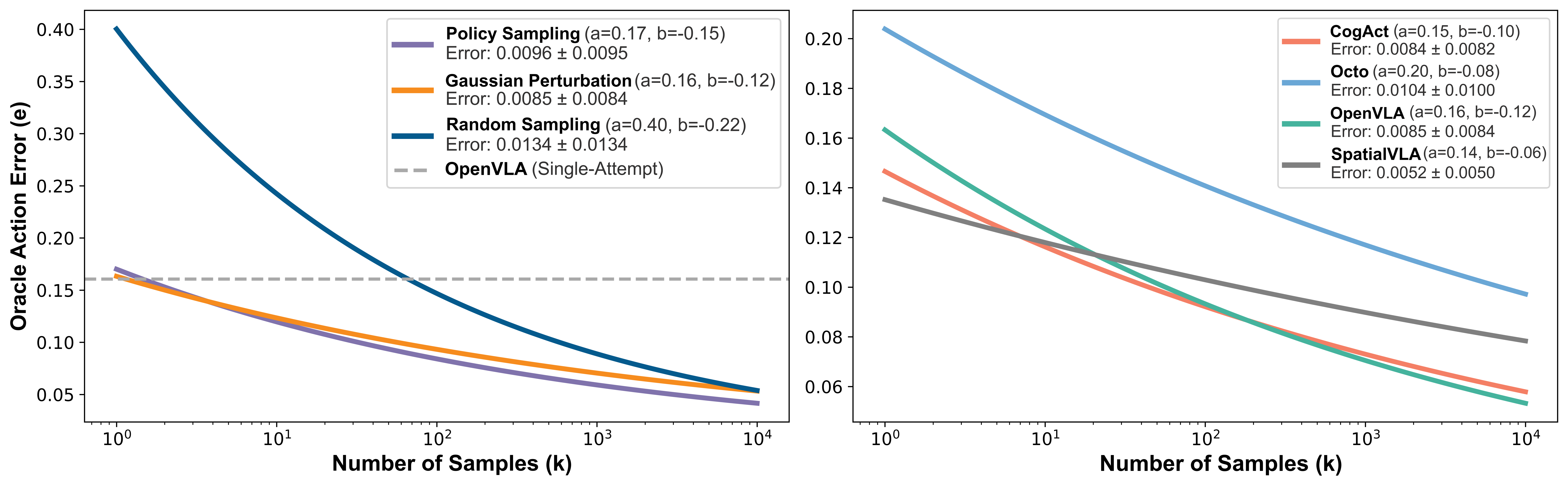}
    \caption{\textbf{Inference-Time Scaling Law:}
    We observe that action error consistently decreases as we scale the number of generated actions across multiple sampling approaches, assuming the presence of an oracle verifier. Repeatedly sampling actions from robot policies, applying Gaussian perturbation to a few sampled actions, and even random sampling all outperform single-attempt OpenVLA. We also find that the relationship between action error and the number of samples generated through Gaussian perturbation follows an approximate power law across a range of VLA models, including CogACT, Octo, OpenVLA, and SpatialVLA. For power law fitting, we model the logarithm of action error $e$ as a function of the number of samples $k$: $\log(e) \approx \log(a) + b \cdot \log(k)$, where $a$ and $b$ are fitted model parameters.
    }
    \label{monkeyplot}
    \vspace{-0.2in}
\end{figure*}

We answer this question in two parts. First, we systematically investigate the benefits of scaling test-time compute in the domain of static manipulation tasks, using off-the-shelf generalist VLA models as base policies. Through our experiments, we find that the relationship between action error and the number of generated samples follows an exponentiated power law across a range of VLAs, demonstrating the existence of inference-time scaling laws. This finding aligns with the power-law scaling~\cite{schaeffer2025largelanguagemonkeyspower, brown2024large} observed in LLMs and suggests that, when paired with a robust verifier, repeated sampling can significantly boost the performance of any off-the-shelf VLA model. Interestingly, different sampling techniques—repeatedly sampling actions from VLAs, Gaussian perturbation applied to a few actions, and random sampling—exhibit a similar scaling pattern. Among these, we find Gaussian perturbation to be the most cost-effective approach and it is therefore adopted in deployment. To our knowledge, our work is the first to characterize inference-time scaling laws for VLAs.

Second, we investigate whether capitalizing on these scaling laws with a learned action verifier can improve policy robustness, guided by the intuition from classic complexity theory that verifying proposals is often easier than generating a solution to a task. To do so, we present a preference-based learning recipe to automatically curate synthetic action comparisons for large-scale imitation learning datasets and use it to train a 7B VLM-based action verifier. Our results show that increasing the synthetic preference dataset size leads to consistent performance improvements. We then introduce our test-time scaling framework, RoboMonkey. During deployment, RoboMonkey samples a small batch of actions from a VLA, applies Gaussian perturbation and majority voting to construct an action proposal distribution, and then uses the fine-tuned VLM-based verifier to select the optimal action. Through extensive evaluations, we demonstrate that pairing existing VLAs with RoboMonkey substantially enhances their precision and robustness. 

The contributions of this paper are summarized as follows: 
\begin{enumerate}
\item  We propose efficient methods for action sampling, and demonstrate that the relationship between action error and the number of samples follows an approximate power law across a range of VLAs.
\item  We present a scalable pipeline for automatically generating synthetic action preferences along with a method for training a VLM-based action verifier.
\item We show that our test-time scaling framework significantly enhances VLA performance, achieving a 25\% absolute improvement in real-world out-of-distribution tasks and 9\% on in-distribution SIMPLER environments.
\item We demonstrate that fine-tuning both VLAs and action verifiers yields a 7\% performance increase compared to fine-tuning VLAs alone on the LIBERO-Long benchmark.
\end{enumerate}

\vspace{0.1in}
\section{Preliminaries}\label{sec:prelim}

We consider a Markov Decision Process $M = ( \mathcal{S},\  \mathcal{A},\ P,\ R)$, where $ \mathcal{S} \subseteq \mathbb{R}^n$ and $ \mathcal{A} \subseteq \mathbb{R}^m$ denote the robot's state and action spaces, respectively. In this work, both the state and action spaces are 7-dimensional vector spaces corresponding to the robot's end effector pose and characterized by three translational states $(x,\ y,\ z) \in \mathbb{R}^3$ and three rotational states $(u,\ v,\ w) \in\mathbb{R}^3$, while the last dimension corresponds to a binary state $g\in\{0, \ 1\}$ indicating whether the end effector gripper is open. $a_t = [\Delta x_t,\ \Delta y_t,\ \Delta z_t,\ \Delta u_t,\ \Delta v_t,\ \Delta w_t,\ g_t]'$ indicates the desired magnitude and direction to augment each state variable at time step $t$. Further, $P(s^\prime \mid s,\ a) \in [0,\ 1]$ represents the robot's non-deterministic transition dynamics from the current state $s \in \mathcal{S}$ with action $a \in \mathcal{A}$ to the candidate state $s^\prime \in \mathcal{S}$, and $R:\ \mathcal{S} \times \mathcal{A} \times \mathcal{I} \rightarrow \mathbb{R}$ provides the reward for choosing action $a \in \mathcal{A}$ at state $s \in \mathcal{S}$ under the language instruction $I \in \mathcal{I}$, where $\mathcal{I}$ is the set of possible instructions. Our framework assumes access to a language-conditioned robot policy $\pi_\theta: \mathcal{S} \times \mathcal{I} \rightarrow \mathcal{A}$, parameterized by $\theta \in \mathbb{R}^{|\theta|}$, from which we can sample multiple actions given the state at timestep $t$ and the language instruction $I \in \mathcal{I}$. Additionally, we assume access to a dataset of $N_D$ expert demonstrations performing a suite of manipulation tasks: $\mathcal{D} = \{(\tau^i,\ I^i)\}_{i=1}^{N_D}$, where each demonstration constitutes a valid trajectory $\tau^i = (s_0^i, a_0^i, \ldots, s_T^i)$ under the transition dynamics to time horizon $T \in \mathbb{N}_+$. We further curate an auxiliary dataset $\mathcal{D}_{\text{buf}} \subset \mathcal{D}$ comprising of tuples $(s_t,\ a_t^*,\ I)$, where $a_t^* \in \mathcal{A}$ is the ground-truth action taken by the expert at state $s_t$ under instruction $I$. We assume that the generalist policy $\pi_\theta$ is fine-tuned to imitate expert demonstrations on this dataset by minimizing the standard imitation learning objective as
$
\mathcal{L}(\theta;\ \mathcal{D}) = -\mathbb{E}_{(s_t^j,\ a_t^j,\ I^j) \sim \mathcal{D}} \left[ 
\log \ \pi_{\theta}(a_t^j \mid s_t^j,\ I^j)\right].$


\section{Inference-Time Scaling Law}\label{sec:scaling}

The relationship between a VLA's action error and its training compute~\cite{sartor2025neuralscalinglawsrobotics, lin2025datascalinglawsimitation} has been well-documented. However, the potential benefits of scaling test-time compute for VLAs remain largely underexplored. To bridge this gap, we conduct a detailed analysis on the Bridge V2 Dataset~\cite{walke2023bridgedata}, examining the relationship between the number of generated samples and action error.

Concretely, we uniformly sample 1,000 $(s,\ a^*,\ I)$ tuples from our auxiliary dataset $\mathcal{D}_{\text{buf}}$. For each tuple, we generate 10,000 actions using various sampling approaches and compute the Normalized Root Mean Squared Error (RMSE) between the ground-truth action $a^*$ and each sampled actions $\{a_1,\ a_2,\ \ldots,\ a_{10,000}\}$.

We evaluate three sampling approaches: \textbf{Random sampling}: candidate actions are generated by uniformly sampling discrete action tokens for each dimension based on the scheme introduced by Brohan et al.~\cite{brohan2023rt} \textbf{Policy sampling}: actions are repeatedly sampled from a robot policy $\pi_\theta(a \mid s,\ I)$ with a positive temperature. \textbf{Gaussian perturbation}: sampling only 4 actions from a robot policy $\pi_\theta(a \mid s,\ I)$, then fitting a Gaussian distribution from which all candidate actions are drawn (see Section \ref{sec:robomonkey} for details).

The result is shown in the left plot of Figure~\ref{monkeyplot}. Assuming the presence of an oracle verifier that always selects the action with the lowest RMSE, we observed that as we scale the number of generated samples, the action error consistently decreases across all sampling methods. Our key findings are: (1) sampling more than 100 actions uniformly at random outperforms greedy decoding using OpenVLA; (2) using policy sampling to repeatedly generate actions from a VLA consistently yields the lowest action error; and (3) Gaussian perturbation achieves nearly identical performance compared to policy sampling while being computationally more efficient. A comprehensive latency analysis is provided in Section~\ref{sec:practical}. 

The right plot of Figure~\ref{monkeyplot} demonstrates that scaling the number of generated samples with Gaussian perturbation is effective across various generalist robot policies, including CogACT, Octo, OpenVLA, and SpatialVLA~\cite{li2024cogactfoundationalvisionlanguageactionmodel, team2024octo, kim2024openvla, qu2025spatialvlaexploringspatialrepresentations}. We find that the relationship between action error and the number of samples often follows an exponentiated power law. Specifically, for OpenVLA, the RMSE decreases by 59.3\% when sampling 10,000 actions. Overall, we offer a new perspective on how we might approach general robot foundation models. Rather than framing robot control purely as a generation problem, our results suggest that viewing it through the lens of verification—generating diverse candidates and verifying them—can substantially improve performance. We hope our findings will motivate and guide the development of scalable action verifiers for robot policies.

\clearpage

\vspace{-0.1in}
\section{Proposed Approach: RoboMonkey}
\vspace{-0.12in}
\begin{figure*}[h!]
    \centering    \includegraphics[width=1.0\textwidth]{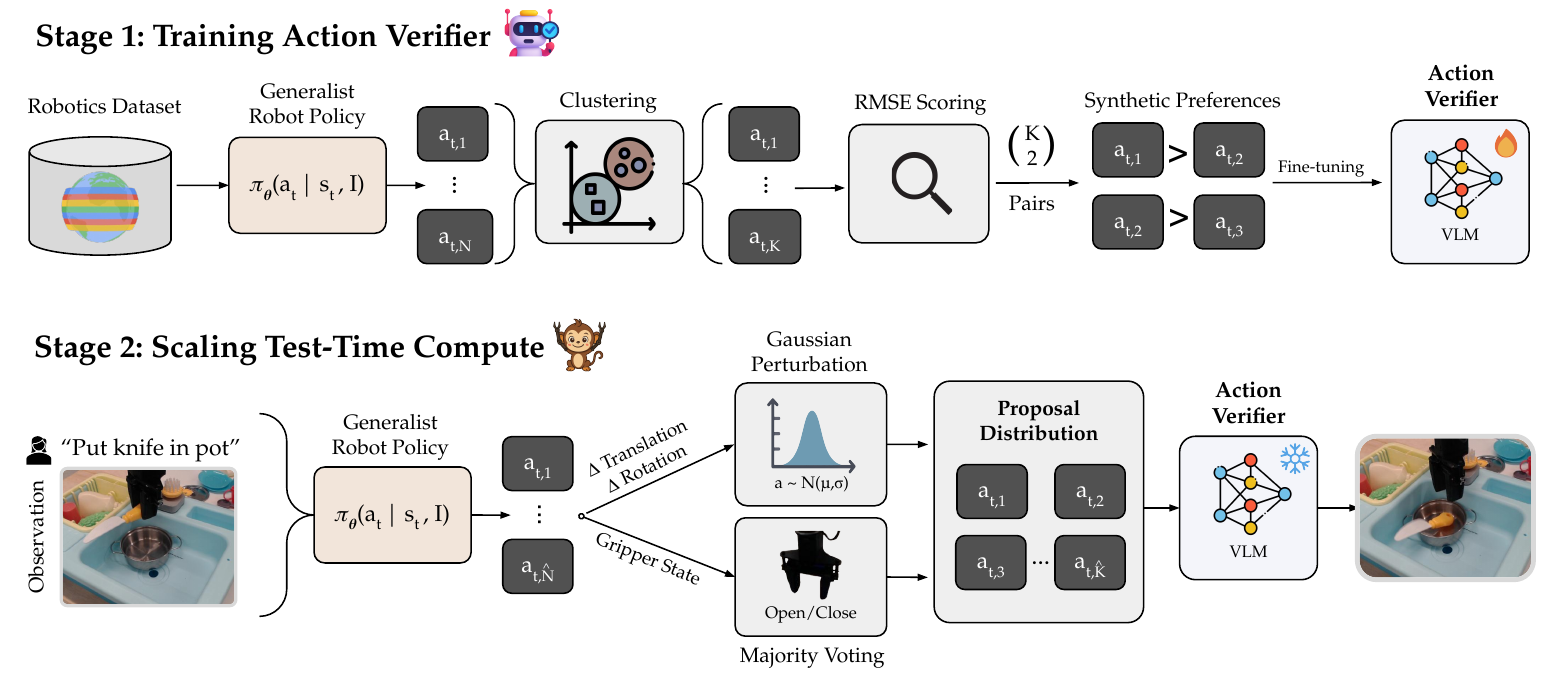}
    \vspace{-0.1in}
    \caption{
    \textbf{Stage 1: Training the Action Verifier.} Given an imitation learning dataset, we sample $N$ candidate actions per state from a generalist robot policy, and apply clustering to reduce them to $K$ representative actions. We construct $\binom{K}{2}$ synthetic action comparisons and assign preferences based on the RMSE between each sampled action and the ground-truth action. This synthetic preference dataset is then used to fine-tune a VLM-based action verifier.
    \textbf{Stage 2: Scaling Test-Time Compute.} At deployment, we sample $\hat{N}$ initial actions from the generalist robot policy based on the given task instruction and observation. We fit a Gaussian distribution $\mathcal{N}(\mu, \sigma)$ to the translation and rotation components $(\Delta x,\ \Delta y,\ \Delta z,\ \Delta u,\ \Delta v,\ \Delta w)$ of these actions, as introduced in \Cref{sec:prelim}, and use majority voting to determine the gripper state. This creates an action proposal distribution from which we can efficiently sample candidate actions with negligible overhead. Finally, we use the fine-tuned VLM-based verifier to evaluate these $\hat{K}$ candidate actions and select the optimal action.}
    \label{pipeline}
    \vspace{-0.1in}
\end{figure*}

\subsection{Motivation}
After establishing the potential for scaling test-time compute for robotics in Section~\ref{sec:scaling}, we now present RoboMonkey, a framework that leverages a learned action verifier to scale test-time compute. We first describe our method for curating a synthetic action preference dataset, followed by reward modeling and inference-time techniques used within RoboMonkey's generate-then-verify pipeline.

\subsection{Synthetic Data Generation Pipeline}\label{section:synthetic}
In this section, we outline our approach for generating synthetic action comparisons, which leverages an existing demonstration dataset $\mathcal{D}$ to produce action pairs with high-quality preference labels without the need for human annotation. Specifically, for each tuple $(s_t,\ a_t^*,\ I)$ from our auxiliary dataset $\mathcal{D}_{\text{buf}}$, we use a reference robot policy to generate $N$ candidate actions. To ensure diversity among the samples, we apply clustering algorithms, reducing these candidates to $K$ representative actions. Subsequently, we construct $\binom{K}{2}$ pairwise comparisons and compute the RMSE between each sampled action, $\{a_t^1,\ a_t^2,\ \dots,\ a_t^K\}$, and the ground-truth action, $a_t^*$. Then, the ``winning" action, $a_t^W$, and the ``losing" action, $a_t^L$, between any two actions $a_t^i$ and $a_t^j$ are determined as follows:
\[
(a_t^W, \ a_t^L) = 
\begin{cases}
    (a_t^i,\ a_t^j), & \text{if } \text{RMSE}(a_t^i,\ a_t^*) < \text{RMSE}(a_t^j,\ a_t^*), \\[6pt]
    (a_t^j,\ a_t^i), & \text{otherwise.}
\end{cases}
\]


We use this procedure to instantiate our action preference dataset $\mathcal{D}_{\text{comp}}$ consisting of tuples $(a_t^W,\ a_t^L,\ a_t^*,\ s_t,\ I)$. Following Ouyang et.al~\cite{ouyang2022traininglanguagemodelsfollow}, we take all $\binom{K}{2}$ pairwise comparisons from identical initial conditions $(s_t,\ I)$ and group these together as a single batch for training.

\subsection{Reward Modeling} \label{sec:rm}
The loss function for training the reward model follows the Bradley-Terry model~\cite{touvron2023llama} with additional modifications to account for preference levels. Formally, we define the ground truth preference level as $\Delta_t^* = \left| \text{RMSE}(a_t^W,\ a_t^*) - \text{RMSE}(a_t^L,\ a_t^*) \right| $ and the predicted preference level from our action verifier $R_{\phi}: A\times S\times\mathcal{I}\rightarrow\mathbb{R} \text{, parameterized by } \phi \in \mathbb{R}^{|\phi|}$, as $\hat{\Delta}_t = \left| R_\phi(a_t^W,\ s_t,\ I) - R_\phi(a_t^L,\ s_t,\ I)\right|$. These components are then integrated into our loss function for training the reward model:
\[
\mathcal{L}(\phi;\ \mathcal{D}_\text{comp}) = -\mathbb{E}_{(a_t^W,\ a_t^L,\ a_t^*,\ s_t,\ I) \sim D_{\text{comp}}} 
\left[ 
\log\ \sigma \left( R_\phi(a_t^W,\ s_t,\ I) - R_\phi(a_t^L,\ s_t,\ I)
- \alpha \left\| \Delta_t^* - \hat{\Delta}_t \right\|^2_2
\right) \right],
\]
where $\sigma : \mathbb{R} \rightarrow [0,\ 1]$ is the sigmoid function and $\alpha\in\mathbb{R}$ is a hyperparameter to control the magnitude of the preference level. We find that including the margin component $\left\| \Delta_t^* - \hat{\Delta}_t \right\|^2_2$ improves the accuracy, particularly when distinguishing between clearly different actions. For more detailed analysis and ablation studies, please refer to Appendices~\ref{E} and~\ref{F}. The action verifier uses LLaVA-7B~\cite{liu2023llava, sun2023aligninglargemultimodalmodels} as the backbone and replaces its final unembedding layer with a reward head. The architecture integrates ViT-Large~\cite{radford2021learningtransferablevisualmodels} as the vision encoder and uses a MLP to map the visual features into the same dimensionality as the word embedding space of the language model.

\subsection{Action Sampling and Verification}\label{sec:robomonkey}

A visualization of the pipeline is shown in Figure~\ref{pipeline}. Formally, at each timestep $t$ during deployment under instruction $I$, RoboMonkey first samples $\hat{N}$ candidate actions from a VLA model $\pi_\theta(a \mid s_t,\ I;\ \mathcal{T})$ with a positive temperature $\mathcal{T}$, yielding a set of candidate actions $\hat{A} = \{\hat{a}_t^1,\ \ldots,\ \hat{a}_t^{\hat{N}}\} \in \mathbb{R}^{m \times \hat{N}}$. Given these samples, we determine the gripper action $g_t$ via majority voting over the discrete gripper component: $g_t = \text{mode}(\{g_t^i\}_{i=1}^{\hat{N}})$. We then fit a Gaussian distribution $\mathcal{N}(\mu_t, \Sigma_t)$ to both the translational components $\{[\Delta \hat{x}_t^i,\ \Delta \hat{y}_t^i,\ \Delta \hat{z}_t^i]'\}_{i=1}^{\hat{N}}$ and rotational components $\{[\Delta \hat{u}_t^i,\ \Delta \hat{v}_t^i,\ \Delta \hat{w}_t^i]'\}_{i=1}^{\hat{N}}$. RoboMonkey then samples $\hat{K}$ new actions from this proposal distribution and appends the fixed gripper state $g_t$ to each, forming a refined action set $\tilde{A} = \{\tilde{a}_t^1,\ \ldots,\ \tilde{a}_t^{\hat{K}}\} \in \mathbb{R}^{m \times \hat{K}}$. Finally, each action $\tilde{a}_t^i$ is scored using our reward model $R_\phi(\tilde{a}_t^i,\ s_t,\ I)$ from which we select the action with the highest reward for execution $a_t = \arg\max_{\tilde{a}_t^i \in \{\tilde{a}_t^1,\ \ldots,\ \tilde{a}_t^{\hat{K}}\}} R_\phi(\tilde{a}_t^i,\ s_t,\ I)$. Below, \Cref{alg:RoboMonkey} presents our detailed test-time scaling pipeline.

\begin{algorithm}
\caption{RoboMonkey Execution}
\label{alg:RoboMonkey}
\KwIn{Generic VLA model $\pi_\theta: \mathcal{S} \times \mathcal{I}\rightarrow A$, reward model $R_\phi: A\times  \mathcal{S} \times\mathcal{I}\rightarrow\mathbb{R}$, initial state $s_0 \in \mathcal{S}$, task instruction $I \in\mathcal{I}$, temperature $\mathcal{T}\in\mathbb{R}_{++}$, time horizon $T\in\mathbb{N}_+$, number of VLA samples $\hat{N}\in\mathbb{N}_+$, number of Gaussian samples $\hat{K}\in\mathbb{N}_+.$}
\For{$t = 0, 1, \ldots, T$}{
    Sample $\hat{A}_t = \{\hat{a}_t^i\}_{i=1}^{\hat{N}} \sim \pi_\theta(a \mid s_t,\ I;\ \mathcal{T})$ \\
    Compute gripper state $g_t \gets \text{mode}(\{\hat{g}_t^i\}_{i=1}^{\hat{N}})$ \\
    Fit Gaussian distribution $\mathcal{N}(\mu_t,\ \Sigma_t)$ on $\{[\Delta \hat{x}_t^i,\ \Delta \hat{y}_t^i,\ \Delta \hat{z}_t^i,\ \Delta \hat{u}_t^i,\ \Delta \hat{v}_t^i,\ \Delta \hat{w}_t^i]'\}_{i=1}^{\hat{N}}$ \\
    Sample $\tilde{A}_t = \{\tilde{a}_t^i\}_{i=1}^{\hat{K}} \sim \mathcal{N}(\mu_t,\ \Sigma_t)$ \\
    Set $\tilde{a}_t^i \gets [\Delta \tilde{x}_t^i,\ \Delta \tilde{y}_t^i,\ \Delta \tilde{z}_t^i,\ \Delta \tilde{u}_t^i,\ \Delta \tilde{v}_t^i,\ \Delta \tilde{w}_t^i,\ g_t]'$ for all $i\in\{1,\ 2,\ \ldots,\ \hat{K}\}$ \\
    Select action $a_t \gets \arg\max_{\tilde{a}_t^i \in \{\tilde{a}_t^1,\ \ldots,\ \tilde{a}_t^{\hat{K}}\}} R_\phi(\tilde{a}_t^i,\ s_t,\ I)$ \\
    Execute $a_t$ and observe $s_{t+1}$ \\
}
\end{algorithm}

\vspace{-0.1in}
\section{Experiments}
The goal of our experiments is to evaluate the effectiveness of scaling test-time compute for generalist robot policies. We evaluate RoboMonkey across both simulated and real-world environments using two different embodiments, across 4 real-robot tasks and 14 simulation tasks.

\begin{figure*}[h!]
    \centering
    \includegraphics[width=1\linewidth]{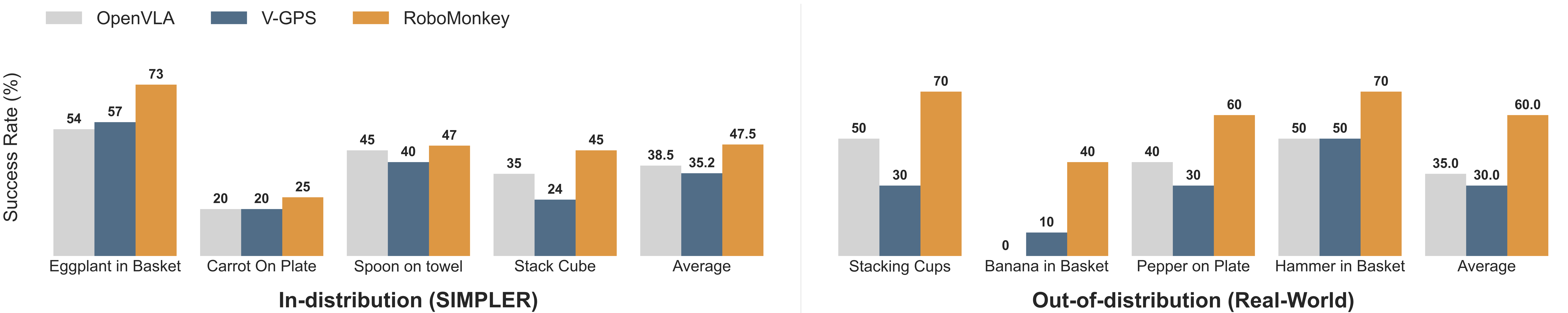}
    \caption{Scaling test-time compute significantly improves the precision and robustness of generalist robot policies across a wide range of manipulation tasks. We observe an 9\% increase in average success rate on in-distribution SIMPLER environments~\cite{li24simpler}, and a 25\% improvement in real-world out-of-distribution experiments using the WidowX robot.}
    \label{fig:mainresult}
\end{figure*}

\subsection{Implementation Details}
We use the Bridge V2 Dataset~\cite{walke2023bridgedata} as our primary training dataset, which includes over 40,000 real-world robotic trajectories collected using the 6-DoF WidowX robot in 24 distinct environments. Following the procedure described in Section~\ref{section:synthetic}, we curated a synthetic action preference dataset consisting of 20 million comparisons. Training was conducted on 8 NVIDIA H100 GPUs with a batch size of 256 using LoRA (r=512, $\alpha$=128). We use OpenVLA as the base model for all experiments. Both RoboMonkey and V-GPS~\cite{nakamoto2025steeringgeneralistsimprovingrobotic} are evaluated by pairing OpenVLA with their respective verifier checkpoints. In real-world evaluation, the system runs at approximately 1.5 Hz on a single NVIDIA H100 GPU and uses a total of 28 GB of GPU memory. For more details about model training and deployment, please refer to Appendix~\ref{B}.
\vspace{0.05in}

\begin{figure*}[h!]
    \centering
\includegraphics[width=1.0\linewidth]{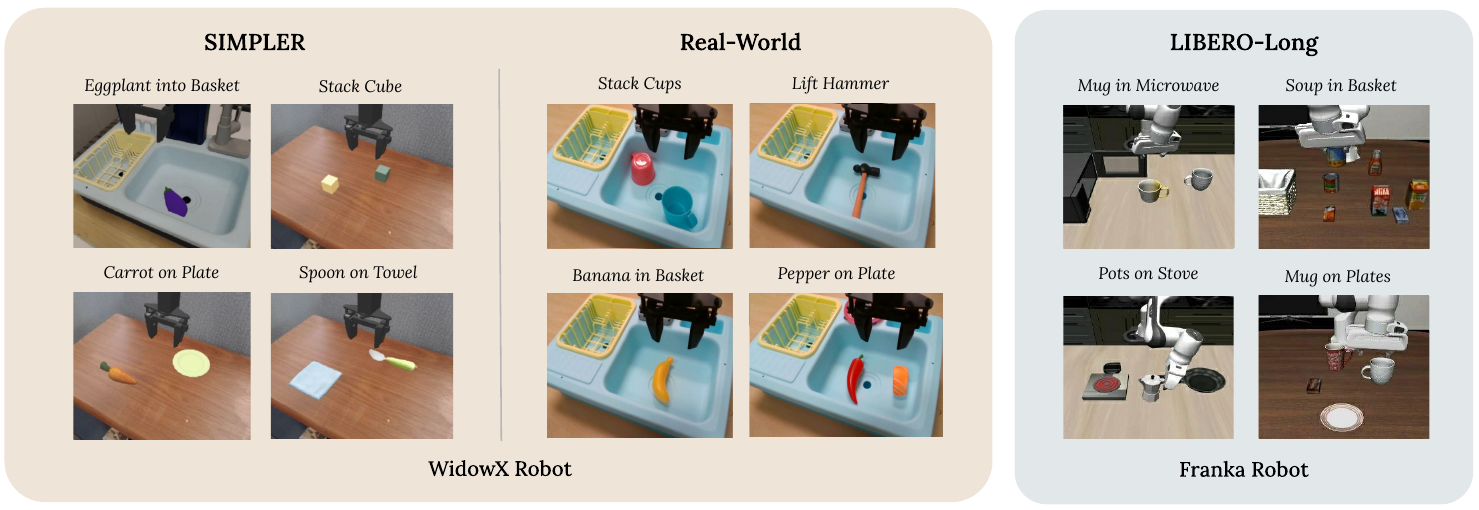}
    \caption{Example tasks across SIMPLER, real-world, and LIBERO environments.}
    \label{task_suites}
    \vspace{-0.1in}
\end{figure*}

\subsection{Can RoboMonkey improve the precision of VLAs on in-distribution tasks?}
\vspace{-0.05in}
We first evaluate our model within the SIMPLER~\cite{li24simpler} environment on in-distribution tasks. This simulation environment is specifically designed to bridge the real-to-sim gap by replicating real-world conditions for WidowX robots. It has demonstrated a strong correlation between performance in SIMPLER and real-world results~\cite{li24simpler}. We evaluate RoboMonkey and other baselines on four tasks: put eggplant in yellow basket, put carrot on plate, put spoon on towel, and stack green block on yellow block.

Figure~\ref{fig:mainresult} presents the evaluation results. RoboMonkey achieves an average success rate of 47.5\%, outperforming OpenVLA by 9\% on average. In the task of placing an eggplant into a basket, RoboMonkey outperforms OpenVLA by 19\%. We observe that the base policy frequently collides with the wall when attempting to move the eggplant toward the basket. Similarly, in the block-stacking task, RoboMonkey surpasses OpenVLA by 10\%. This task requires accurate grasping and precise placement of small objects. Overall, these results highlight that making local refinements to the base policy’s actions can substantially improve precision. Additionally, we observe that pairing V-GPS with OpenVLA results in worse performance than both standalone OpenVLA and RoboMonkey. See Appendices~\ref{A} and~\ref{C} for details on the task setup and the ablation study on action selection.

\subsection{Can RoboMonkey improve the robustness of VLAs on out-of-distribution tasks?} \label{5.3}
For a more comprehensive evaluation, we design a set of real-world manipulation tasks on a physical WidowX-250 S robot to evaluate RoboMonkey in OOD settings. As illustrated in Figure~\ref{task_suites}, these tasks include unseen instructions, objects, distractors, and shapes. We evaluate each approach across 4 task suites with 10 trials each, resulting in a total of 120 rollouts. Figure~\ref{fig:mainresult} compares the performance of RoboMonkey, V-GPS, and OpenVLA across a suite of tasks on the WidowX robot. RoboMonkey consistently outperforms both baselines across diverse tasks, including stacking cups, lifting a hammer, placing banana into a yellow basket, and putting a pepper onto a plate. We find that RoboMonkey effectively mitigates issues of imprecise grasping, task progression failures, and collisions at deployment. Detailed task breakdowns and failure analysis are provided on our project page: \href{https://robomonkey-vla.github.io}{https://robomonkey-vla.github.io}.

Notably, RoboMonkey exhibits substantial improvements on tasks requiring visual and semantic generalization. For example, in the banana-in-basket task, OpenVLA achieved a success rate of 0\%, as it lacks the language and visual grounding to differentiate between two yellow objects (banana and yellow basket), thus making no progress in completing the task. Furthermore, RoboMonkey achieves over 20\% higher success rates on fine-grained manipulation tasks such as cup stacking and hammer lifting. These tasks require precise reasoning about grasp points, particularly on novel objects and shapes. Overall, RoboMonkey achieved an average success rate of 60\%, compared to 35\% from OpenVLA and 30\% from V-GPS, indicating that our action verifier is significantly less sensitive to distribution shifts than the base policy. As such, these results underscore RoboMonkey’s effectiveness in improving the robustness and generalization in OOD scenarios.

\begin{figure*}[h!]
    \centering
\includegraphics[width=1\linewidth]{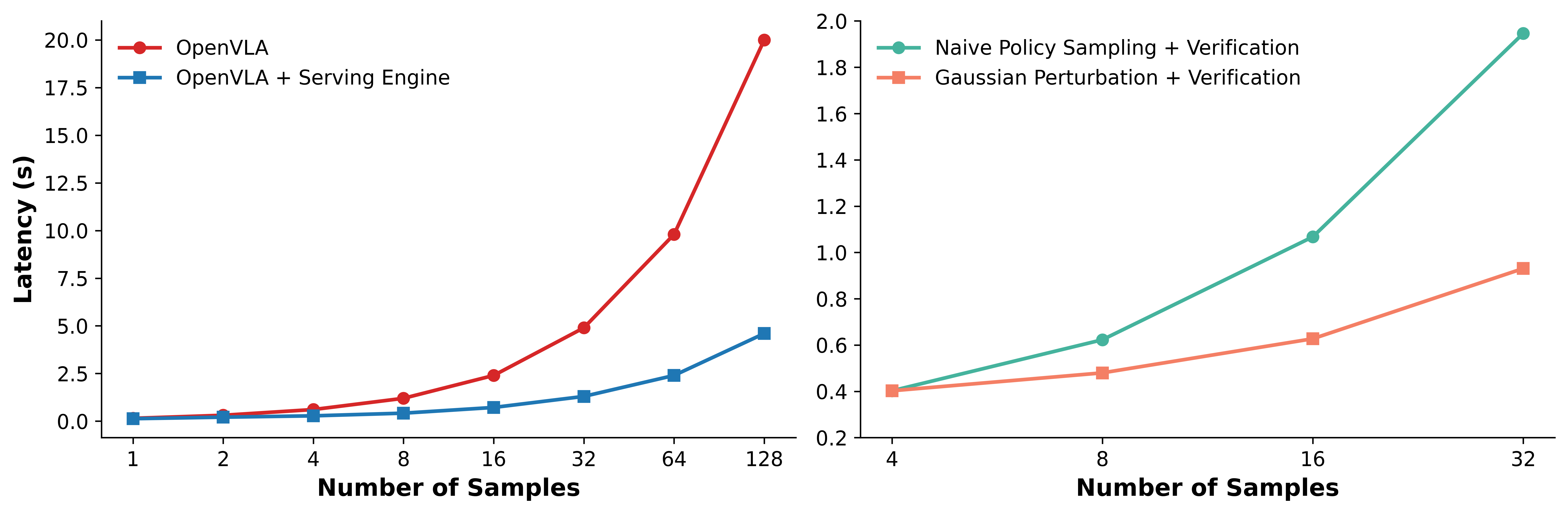}
    \caption{\textbf{Left:} Repeated sampling can exploit KV Cache optimizations and batch processing to achieve higher throughput than greedy decoding. Therefore, we extended SGLang’s capabilities to properly support OpenVLA. Our optimized implementation substantially outperforms the naive OpenVLA inference pipeline, achieving lower latency and significantly higher throughput across batch sizes. \textbf{Right:} Latency comparison between naive policy sampling and Gaussian perturbation as we scale the number of samples.}
    \label{fig:throughput}
    \vspace{0.07in}
\end{figure*}

\subsection{How does RoboMonkey enable practical deployment for test-time scaling?} \label{sec:practical}
While RoboMonkey introduces additional computational overhead from action sampling and verification, we mitigate these costs with a practical serving solution. Specifically, we implemented a VLA serving engine on top of SGLang~\cite{zheng2024sglangefficientexecutionstructured} to speed up repeated sampling of initial action candidates (see Figure~\ref{fig:throughput}) and employ Gaussian perturbation to efficiently construct an action proposal distribution, as detailed in Section~\ref{sec:robomonkey}. With these optimizations, RoboMonkey can sample and verify 16 candidate actions in approximately 650 ms (or 1.5 Hz), achieving a 41.3\% lower latency compared to naive policy sampling, as shown in Figure~\ref{fig:throughput} (right). Gaussian perturbation proves more efficient because latency scales only with verification cost, whereas naive policy sampling incurs increasing latency from both sampling and verification as the number of candidate actions grows. See Appendix~\ref{H} for a detailed analysis of the trade-off between action error and compute budget.
\begingroup
\renewcommand{\thefootnote}{}
\footnote{SIMPLER results were obtained using two NVIDIA RTX 4090 GPUs, while real-world experiments and latency analysis were conducted on a single NVIDIA H100. LIBERO evaluations used an NVIDIA RTX 6000 Ada.}
\endgroup
\renewcommand{\thefootnote}{\arabic{footnote}}
\clearpage

\subsection{How does scaling the synthetic training dataset impact downstream success rate?}
    \vspace{-0.1in}
\begin{figure*}[h!]
    \centering
\includegraphics[width=0.6\linewidth]{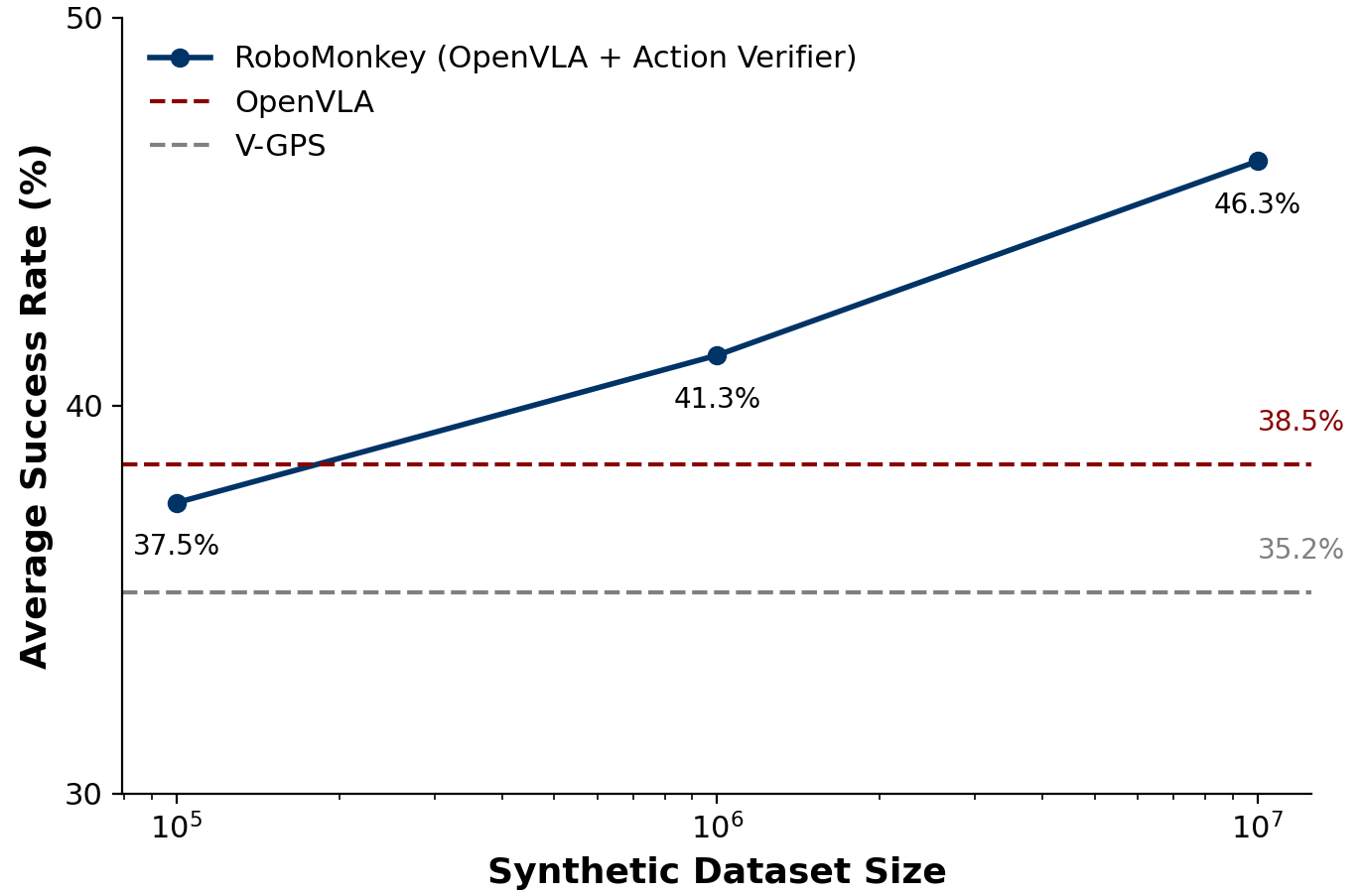}
    \caption{Average success rates across four tasks on SIMPLER as a function of synthetic dataset size. Scaling the dataset size (number of synthetic action comparisons) consistently improves the performance of the RoboMonkey action verifier, leading to higher closed-loop success rates.}
    \label{fig:synthetic}
\end{figure*}

We demonstrate that RoboMonkey's closed-loop success rates on SIMPLER consistently improve as the synthetic dataset size increases, with the average success rate rising from 37.5\% to 46.3\% as shown in Figure ~\ref{fig:synthetic}. With action verifiers trained on over $10^6$ action comparisons, RoboMonkey outperforms both OpenVLA and V-GPS. The improvements are particularly pronounced in fine-grained manipulation tasks like ``Stacking Cube'', where success rates increase from 27\% to 37\% to 42\% as the synthetic dataset scales. We find that the overall task performance grows nearly log-linearly with synthetic dataset size, highlighting the potential of large-scale synthetic data generation for enhancing action verification.

\subsection{Can we effectively fine-tune the action verifier on new robot setup and task?}
\vspace{-0.1in}
\begin{table}[h!]
\centering
\small 
\setlength{\tabcolsep}{5pt} 
\renewcommand{\arraystretch}{0.9} 
\begin{tabular}{lcc}
\toprule
\textbf{Task} & \textbf{OpenVLA} & \textbf{RoboMonkey} \\
\midrule
Soup and Sauce in Basket & 36\% & \textbf{59\%} \\
Cheese and Butter in Basket & 70\% & \textbf{79\%} \\
Turn on Stove and Place Moka & 58\% & \textbf{58\%} \\
Black Bowl in Drawer & 36\% & \textbf{37\%} \\
Mugs on Plates & 42\% & \textbf{55\%} \\
Book in Caddy & 84\% & \textbf{86\%} \\
Mug and Pudding on Plate & 48\% & \textbf{59\%} \\
Soup and Cheese in Basket & 56\% & \textbf{62\%} \\
Moka Pots on Stove & 26\% & \textbf{26\%} \\
Mug in Microwave & 42\% & \textbf{44\%} \\
\midrule
\textbf{Average Success Rate} & 49.8\% & \textbf{56.5\%} \\
\bottomrule
\end{tabular}
\vspace{5pt}
\caption{Comparison of task success rates between OpenVLA and RoboMonkey, both fine-tuned and evaluated on the LIBERO-Long benchmark.}
\label{tab:openvla_comparison_libero90}
\end{table}
\vspace{-4pt}

We further evaluate RoboMonkey's adaptability on a new robot setup. In this section, we present the fine-tuning evaluation of RoboMonkey on the LIBERO-Long benchmark, which consists of long-horizon tasks in simulation. In our experiments, we curated a new action preference dataset using the procedure described in Section~\ref{section:synthetic} for fine-tuning the action verifier. The OpenVLA-LIBERO checkpoint used for comparison was trained via behavioral cloning on successful demonstrations. All methods were evaluated across 500 trials. Table~\ref{tab:openvla_comparison_libero90} presents the results and we observe that RoboMonkey can be effectively adapted to tasks in the LIBERO environments. We find that fine-tuning both OpenVLA and action verifier results in 6.7\% improvement in average success rate compared to simply fine-tuning OpenVLA on LIBERO-Long.

\vspace{-1pt}
\section{Discussion and Limitations}
\vspace{-1pt}

In this paper, we presented RoboMonkey, a novel test-time scaling framework that enhances the precision and robustness of Vision-Language-Action (VLA) models. RoboMonkey achieves significant performance improvements across both in-distribution and out-of-distribution tasks, as well as on new robot setups. Our findings demonstrate that scaling test-time compute through a generate-and-verify paradigm provides a practical and effective path towards building general-purpose robotics foundation models. The current RoboMonkey framework has several limitations that we leave for future work:
\vspace{-1pt}

\paragraph{Computational Overhead:} Since it requires sampling multiple candidate actions from a VLA and employs a separate VLM-based action verifier, it incurs increased computational overhead during deployment. Although we mitigate these costs with a practical serving solution—using a VLA serving engine and Gaussian perturbation—the framework may still be less suitable for tasks requiring high-frequency control. Future work could explore more efficient model architectures for action verification and apply system-level optimizations to further reduce the memory footprint and latency of scaling test-time compute.
\vspace{-1pt}

\paragraph{Scaling Synthetic Datasets:} Our results show that increasing the size of the synthetic training dataset consistently improves downstream robot performance. However, due to compute constraints and the high cost of fine-tuning, we limited our experiments to 20 million synthetic action comparisons on the Bridge V2 dataset. Scaling synthetic data generation to larger robotics datasets across embodiments, tasks, and environments is a promising direction for future exploration.
\vspace{-1pt}

\paragraph{Evaluation Scope:} While our experiments focused on two commonly used robotic arms—WidowX 250S and Franka—future work should evaluate RoboMonkey across a broader range of embodiments.
\vspace{-1pt}
\vspace{-1pt}

\section{Related Work}
\vspace{-1pt}

\textbf{Vision Language Action Models}: Recent advancements in robotics have seen a shift toward training multi-task generalist robot policies~\cite{jiang2022vima} on large robotics datasets~\cite{walke2023bridgedata, fang2023rh20t} collected on diverse scenes and robot embodiments. In this landscape, several robotics foundation models have emerged. $\pi_0$, OpenVLA, PaLM-E, and RT-X~\cite{black2024pi0visionlanguageactionflowmodel, kim2024openvla, driess2023palmeembodiedmultimodallanguage, brohan2022rt, brohan2023rt} have demonstrated strong generalization capabilities by combining Transformer architectures~\cite{vaswani2017attention} or diffusion policies~\cite{chi2023diffusion} with imitation learning. While these generalist policies demonstrate out-of-the-box capabilities for controlling robots, they may still fail due to distribution shift and compounding prediction errors. Our evaluation demonstrates that RoboMonkey significantly improves the robustness and generalization of these generalist policies at deployment.
\vspace{-1pt}

\textbf{Out-of-Distribution Robustness}: The challenge of learning-based systems performing unreliably on data that differs from their training distribution is documented across robotics literature~\cite{sinha2023systemlevelviewoutofdistributiondata, agia2024unpackingfailuremodesgenerative, sinha2024realtimeanomalydetectionreactive}. Researchers have approached this challenge through various methodologies, including robust training and adapting models to varying environmental distribution shifts~\cite{sinha2023systemlevelviewoutofdistributiondata, zhu2025nerfaugdataaugmentationrobotics, mitrano2022dataaugmentationmanipulation}. A significant breakthrough came with the emergence of Foundation Models (FMs). Recently, FM is widely adopted in robotics. For instance, several prior works~\cite{liang2023codepolicieslanguagemodel,mu2023embodiedgptvisionlanguagepretrainingembodied, shi2025hirobotopenendedinstruction} explore employing VLMs to generate sequences of high-level action plans, which are then executed through low-level policy. In contrast, rather than using them primarily for hierarchical planning, RoboMonkey and several concurrent works~\cite{wu2025foresightforethoughtvlminthelooppolicy, wang2025inferencetimepolicysteeringhuman} employ FMs as action verifiers that evaluate the low-level actions generated by robot policies. 
\vspace{-1pt}

\textbf{Repeated Sampling}: The methodology of applying additional computation at test time has demonstrated remarkable success across various domains. For LLMs, repeated sampling has proven effective in enhancing performance across diverse tasks, including mathematical problem-solving, coding, and text summarization~\cite{chen2024alphamath, brown2024large, ehrlich2025codemonkeysscalingtesttimecompute}. In robotics, V-GPS~\cite{nakamoto2025steeringgeneralistsimprovingrobotic} adopts a related strategy by training a value function with offline RL to re-rank candidate actions, selecting those that lead to better outcomes. RoboMonkey introduces a more scalable data curation pipeline and model architecture for training the action verifier. Our experimental results show that pairing existing VLAs with our verifier substantially improves both task performance and generalization compared to prior verifier-based approaches. Instead of relying on naive action sampling from robot policies, RoboMonkey uses Gaussian perturbation to efficiently generate diverse candidate actions and integrates inference-time techniques such as majority voting to guide the verification process.
\vspace{-1pt}


\acknowledgments{This work was supported by DARPA and the
National Aeronautics and Space Administration under the University Leadership Initiative program.}

\bibliography{example}  

\clearpage
\appendix
\clearpage

\section{Evaluation Tasks} \label{A}
As described in Section~\ref{5.3} and illustrated in Figure~\ref{task_suites}, we evaluate RoboMonkey, OpenVLA, and V-GPS on a physical WidowX-250~S robot across four out-of-distribution (OOD) generalization tasks. Real-world evaluations inherently introduce distribution shifts, as we cannot exactly replicate the Bridge V2 setup. In particular, slight variations in camera placement, robot positioning, lighting conditions, and background are unavoidable. We describe the four representative generalization tasks as follows:

\begin{itemize}
    \item \textbf{Stack Blue Cup on Pink Cup}: The goal is to grasp the blue cup and stack it on top of the pink cup. The language instruction of this task was not included in the Bridge V2 dataset.

    \item \textbf{Put Hammer into Yellow Basket}: The robot must lift the hammer and place it inside a yellow basket. Importantly, the hammer represents an unseen object with a novel shape not present in any Bridge V2 demonstration.

    \item \textbf{Put Pepper onto Plate}: This language grounding task requires the robot to identify and approach the pepper while ignoring distractors (e.g., sushi). The robot must then differentiate between the yellow basket and a plate before correctly placing the pepper.

    \item \textbf{Put Banana into Yellow Basket}: The objective of this task is to place a banana from the sink into the yellow basket. This task presents a particular challenge as the system must differentiate between two yellow objects (banana and yellow basket), requiring visual and language grounding to complete the task successfully.
\end{itemize}

\begin{figure*}[h!]
    \centering
\includegraphics[width=1.0\linewidth]{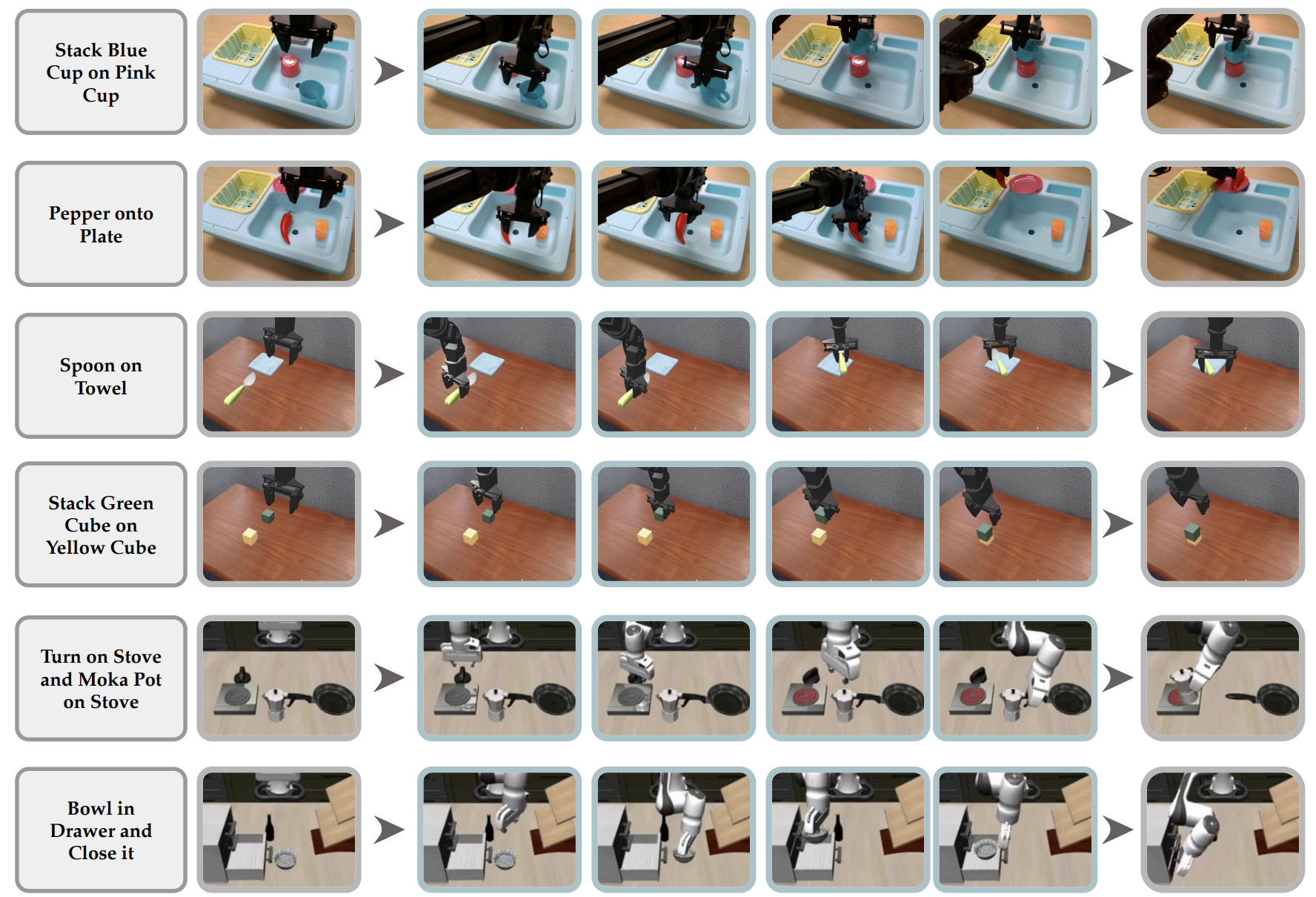}
    \caption{Representative task executions in real-world, SIMPLER, and LIBERO environments.}
    \label{task_exec}
\end{figure*}

For details on the task setup for in-distribution and fine-tuning evaluation, please refer to the SIMPLER~\cite{li24simpler} and LIBERO-LONG~\cite{liu2023libero} benchmark. We include task execution examples in Figure~\ref{task_exec}.

\clearpage
\section{Implementation Details} \label{B}
\subsection{Training}
Our action verifier uses LLaVA-7B~\cite{liu2023llava, sun2023aligninglargemultimodalmodels} as the backbone and replaces its final unembedding layer with a reward head. The architecture integrates ViT-Large~\cite{radford2021learningtransferablevisualmodels} as the vision encoder and uses a MLP to map the visual features into the same dimensionality as the word embedding space of the language model. We discretize each dimension of a continuous action into 256 bins, following Brohan et al.~\cite{brohan2022rt}, and overwrite the 256 least frequent tokens in the LLaMA tokenizer with these discrete action tokens.

Training was conducted on 8 NVIDIA H100 GPUs using LoRA (rank=512, $\alpha=128$), and our codebase builds on top of LLaVA-RLHF~\cite{sun2023aligninglargemultimodalmodels}. We use a batch size of 256 and train on a synthetic preference dataset comprising 20 million comparisons derived from the Bridge V2 dataset. The model is trained using the Adam optimizer with a learning rate of 2e-5. Training is conducted for a single epoch. A margin weight of 0.1 is applied to the modified Bradley-Terry loss. Example prompt to action verifier is shown below:
\begin{quote}
\begin{verbatim}
USER: <image> shows the current observation from the robot's wrist-mounted camera. The robot manipulation arm is attempting to [instruction].
What action should the robot take to effectively accomplish the task?
ASSISTANT: The robot should take the action [Discrete Action Tokens]
USER: Please evaluate the quality of the robot action.
ASSISTANT: The quality score of the robot action is
\end{verbatim}
\end{quote}

\subsection{Deployment}
For real-world evaluation, we first sample 5 initial actions from OpenVLA with temperature 1.0. We fit a Gaussian distribution $\mathcal{N}(\mu, \sigma)$ to the translation and rotation components, and use majority voting to determine the gripper state. This creates an action proposal distribution from which we sample 16 candidate actions. We then use the fine-tuned VLM-based verifier to select the optimal action for execution. We conduct 10 trials per task and report the average success rate. In simulation, we vary the number of initial action samples $\hat{N} \in \{5, 9\}$ and the number of augmented samples $\hat{K} \in \{8, 16, 32\}$. We report the best results for each task. All simulated experiments are conducted using a machine equipped with two NVIDIA RTX 4090 GPUs over three random seeds.

\subsection{Baselines}
We use the publicly released OpenVLA checkpoint from \url{https://huggingface.co/openvla/openvla-7b}, and the V-GPS value function checkpoint from \url{https://github.com/nakamotoo/V-GPS}. In simulation, we follow the evaluation procedure outlined in the V-GPS implementation, sweeping over the number of samples $\{10, 50\}$ and softmax temperatures $\{0, 0.1, 1.0\}$, and report the best result for each task. For real-world evaluations, we fix the number of samples to 10 and the temperature to 1.0. It is worth noting that we use our VLA serving engine to enable efficient batch inference for all experiments.

\section{Ablation Over Action Selection Methods and Number of Samples} \label{C}
\label{appendix:ablation_selection}
\begin{figure*}[h!]
    \centering
    \includegraphics[width=0.6\linewidth]{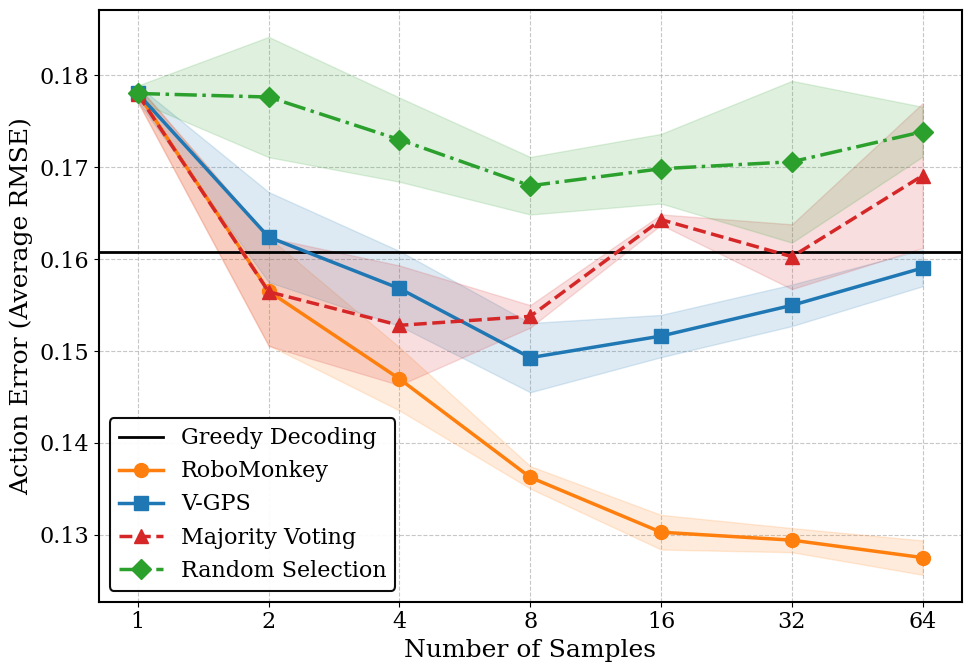}
    \caption{Comparison of action error (average RMSE) across different selection methods as the number of generated samples increases. RoboMonkey consistently outperforms other baselines and scales effectively with additional compute.}
    \label{fig:action_error}
\end{figure*}

To evaluate the effectiveness of our verifier, we adopt a setup similar to that described in Section~\ref{sec:scaling}. Specifically, we uniformly sample 1,000 \((s, a^*, I)\) tuples from the auxiliary dataset \(\mathcal{D}_{\text{buf}}\), curated from the BridgeV2~\cite{walke2023bridgedata}. For each tuple, we generate 64 candidate actions using the reference policy, OpenVLA, and apply various selection techniques—including RoboMonkey, V-GPS, majority voting, and random selection—to identify the optimal action among the samples. We report the normalized RMSE between the ground-truth action and the selected action for each method. As shown in Figure~\ref{fig:action_error}, RoboMonkey consistently achieves the lowest action error across different sample sizes. When generating 64 samples, RoboMonkey reduces the action error by 21\% relative to the greedy decoding baseline, highlighting the effectiveness of our verifier in improving action precision.
While prior work such as V-GPS trained with offline RL also improves over greedy decoding, its value function achieves only a 6\% reduction in action error. Furthermore, we observe that increasing the number of samples leads to exploitation of the V-GPS value function, resulting in performance degradation when sampling more than 8 actions. In contrast, RoboMonkey remains robust to reward hacking and demonstrates scalability with increased test-time compute.

\section{Ablation Over Generalist Robot Policies} \label{D}

We conducted additional experiments to ablate the performance of RoboMonkey when paired with different VLA models. For models that generate action chunks, we apply temporal ensembling and discretize the outputs using the scheme introduced by Brohan et al.~\cite{brohan2023rt} to enable scoring by the action verifier. Following a similar evaluation setup to Appendix \ref{C}, our ablation considers three generalist robot policies: CogACT, Octo, and SpatialVLA.

\begin{figure*}[h!]
    \centering
    \includegraphics[width=1.0\linewidth]{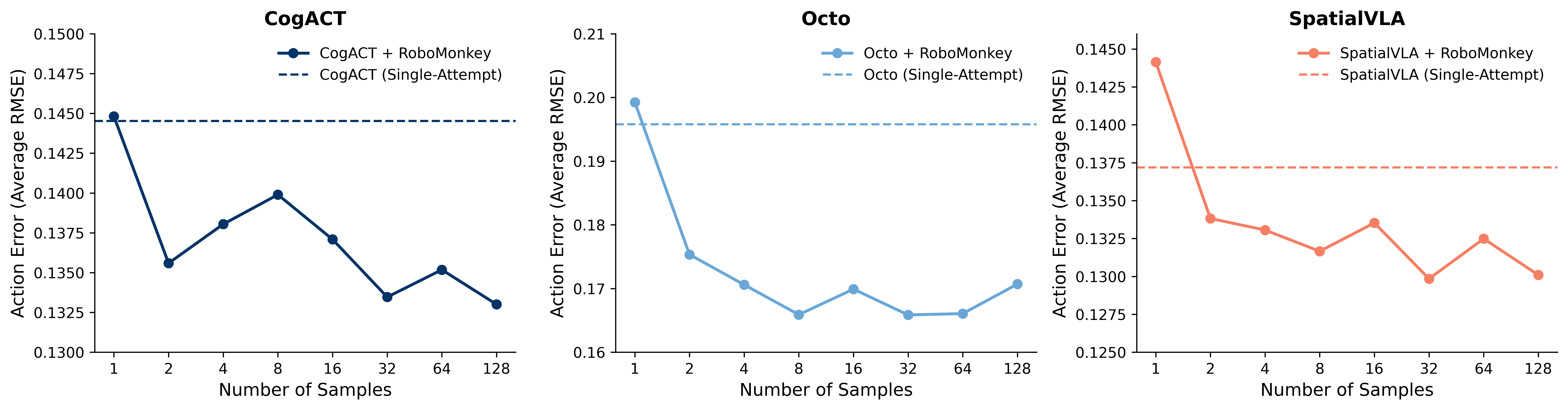}
    \caption{Effect of scaling test-time compute with RoboMonkey across different generalist robot policies. Action error (average RMSE) decreases as the number of samples increases. Dashed lines denote the action error of each base policy when only generating a single action.}
    \label{fig:generalist_ablation}
\end{figure*}

\noindent
\textbf{CogACT} is a 7B-parameter VLA model with a modular architecture that separates cognitive reasoning from motor control. Built on top of the Prismatic VLM (DINOv2 + SigLIP for vision and LLaMA-2 for language), CogACT introduces a specialized action module implemented as a diffusion transformer. We use the CogACT-base variant in our experiments. Pairing RoboMonkey with CogACT achieves RMSE of 0.133, reflecting an 8\% reduction from its single-attempt baseline of 0.145.

\noindent
\textbf{Octo} is a transformer-based generalist policy trained on 800K demonstrations from the Open X-Embodiment (OXE) dataset. The policy includes a CNN encoder and a ViT-style transformer backbone with a diffusion-based action head that predicts action sequences. We use Octo-small for evaluation. Integrating RoboMonkey with Octo achieves RMSE of 0.166, representing a 15.3\% reduction from its greedy baseline (0.196).

\noindent
\textbf{SpatialVLA} is a 3.5B-parameter spatially grounded VLA model trained on 1.1M robot episodes from OXE and RH20T. The model uses Ego3D Position Encoding to integrate 3D spatial context from depth estimates into visual features. It is pre-trained on a PaLI-Gemma-2 backbone. For deployment, we use a temperature of 0.5 for sampling. SpatialVLA + RoboMonkey achieves RMSE of 0.1298, a 5.3\% reduction from its baseline of 0.137.

\vspace{-0.1in}
\section{Ablation on Margin for Reward Modeling} \label{E}
\vspace{-0.1in}

\begin{table}[h!]
\centering
\begin{tabular}{cccc}
\toprule
$\alpha$ & Precision & Recall & F1 Score \\
\midrule
0   & 0.81  & 0.85  & 0.83 \\
0.1 & \textbf{0.84}  & \textbf{0.87}  & \textbf{0.85} \\
1.0 & 0.79  & 0.83  & 0.81 \\
\bottomrule
\end{tabular}
\vspace{0.05in}
\caption{Comparison of action verifier performance across different margin weights $\alpha$ in the loss function. We find that incorporating a small margin ($\alpha = 0.1$) improves precision, recall, and F1 score}
\label{tab:margin_ablation}
\end{table}

As illustrated in Section~\ref{sec:rm}, the loss function for training the action verifier follows the Bradley-Terry model~\cite{touvron2023llama} with an additional margin component to account for preference levels. \[
\mathcal{L}(\phi;\ \mathcal{D}_\text{comp}) = -\mathbb{E}_{(a_t^W,\ a_t^L,\ a_t^*,\ s_t,\ I) \sim D_{\text{comp}}} 
\left[ 
\log\ \sigma \left( R_\phi(a_t^W,\ s_t,\ I) - R_\phi(a_t^L,\ s_t,\ I)
- \alpha \left\| \Delta_t^* - \hat{\Delta}_t \right\|^2_2
\right) \right],
\]
where $\alpha\in\mathbb{R}$ is a hyperparameter to control the magnitude of the preference level. To evaluate the effectiveness of this margin component, we generated 10,000 synthetic action comparison pairs from the Bridge V2 dataset following the procedure outlined in Section~\ref{section:synthetic}. Each action pair consists of distinctly different actions. We trained two variants of the action verifier with margin terms ($\alpha \in {0.1,\ 1.0}$) and compared their performance to a baseline model without a margin term ($\alpha = 0$). Table~\ref{tab:margin_ablation} reports the precision, recall, and F1 score for each setting. The variant with $\alpha = 0.1$ achieved the highest F1 score (0.85), outperforming both the baseline without a margin ($\alpha = 0$, F1 = 0.83) and the large-margin variant ($\alpha = 1.0$, F1 = 0.81). These results suggest that incorporating a margin term can improve the action verifier's performance, but an excessively large margin may negatively impact verification accuracy. Based on this analysis, we adopt the $\alpha = 0.1$ variant for deployment.

\vspace{-0.07in}
\section{Ablation Over Preference-Based Learning} \label{F}
\label{sec:ablation_rmse}
\vspace{-0.07in}

\begin{figure}[h!]
    \centering
    \includegraphics[width=1.0\linewidth]{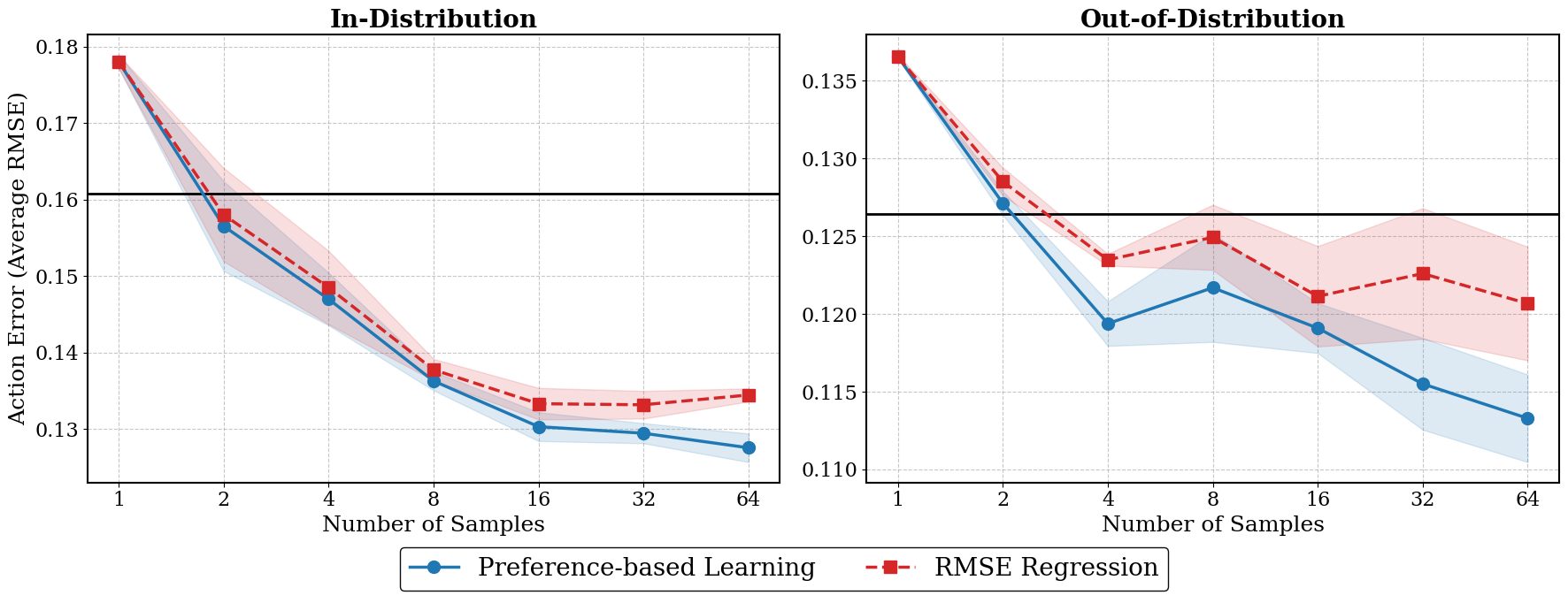}
    \caption{Comparison between preference-based learning and RMSE regression across number of samples. While both perform similarly in-distribution, preference learning generalizes better in OOD settings.}
    \label{fig:rmse_ablation}
\end{figure}
To further understand the effectiveness of our preference-based learning approach, we conduct an ablation comparing our Bradley-Terry objective against a baseline that directly predicts RMSE values. Specifically, we train an alternative action verifier for one epoch to minimize the L2 loss between the predicted and ground-truth RMSE:
\[
\mathcal{L}(\phi; \mathcal{D}_{\text{comp}}) = \mathbb{E}_{(a_t, a_t^*, s_t, I) \sim \mathcal{D}_{\text{comp}}} \left\| R_\phi(a_t, s_t, I) - \text{RMSE}(a_t, a_t^*) \right\|_2^2
\]
where the verifier directly learns to predict the RMSE between any candidate action $a_t$ and the ground-truth action $a_t^*$.

We follow the same setup as described in Section~\ref{sec:scaling} and Appendix~\ref{appendix:ablation_selection} for in-distribution evaluation using the Bridge V2 dataset~\cite{walke2023bridgedata}. For OOD analysis, we sample 1,000 $(s, a^*, I)$ tuples from held-out trajectories with unseen instructions and environments. The error bars represent the standard deviation across three random seeds.

As shown in Figure~\ref{fig:rmse_ablation}, both approaches achieve similar performance on in-distribution environments, with preference-based learning slightly outperforming RMSE regression. However, the performance gap becomes substantial in OOD settings. When sampling 64 candidate actions, preference-based learning achieves a 6\% lower action error compared to RMSE regression. This result reveals a crucial insight: instead of directly regressing RMSE values, preference-based learning teaches the model to make relative comparisons between actions, enabling stronger generalization.

\section{Latency and Throughput Analysis for Sampling and Verification} \label{G}
\begin{table}[h!]
\centering
\begin{tabular}{c|cc|cc|cc}
\toprule
\textbf{Batch} & \multicolumn{2}{c|}{\textbf{OpenVLA}} & \multicolumn{2}{c|}{\textbf{OpenVLA + SGLang}} & \multicolumn{2}{c}{\textbf{Action Verifier}} \\ \textbf{Sizes}
& \textbf{Latency (s)} & \textbf{Throughput} & \textbf{Latency (s)} & \textbf{Throughput} & \textbf{Latency (s)} & \textbf{Throughput} \\
\midrule
1   & 0.15  & 6.5   & 0.13  & 7.6   & 0.092 & 11  \\
2   & 0.31  & 3.3   & 0.21  & 9.6   & 0.099 & 20  \\
4   & 0.61  & 1.6   & 0.28  & 15    & 0.13  & 32  \\
8   & 1.2   & 0.82  & 0.42  & 19    & 0.20  & 39  \\
16  & 2.4   & 0.41  & 0.72  & 22    & 0.35  & 46  \\
32  & 4.9   & 0.20  & 1.3   & 25    & 0.65  & 49  \\
64  & 9.8   & 0.10  & 2.4   & 27    & 1.3   & 50  \\
128 & 20    & 0.050 & 4.6   & 28    & 2.5   & 51  \\
\bottomrule
\end{tabular}
\caption{Latency (seconds) and throughput (samples/second) comparison across batch sizes for OpenVLA, Optimized OpenVLA, and 7B Action Verifier.}
\label{tab:latency_throughput_comparison}
\end{table}

Repeated sampling can exploit KV Cache optimizations and batch processing to achieve higher throughput than greedy decoding. However, most VLA models, including OpenVLA, are built on top of Prismatic VLM and do not support batching~\cite{kim2024openvla}. SGLang provides efficient serving with prefix caching, overhead-free CPU scheduling, and paged attention. Therefore, to make RoboMonkey practical for deployment, we extended SGLang’s capabilities to properly support Prismatic VLM~\cite{kim2024openvla} models, enabling us to achieve higher throughput during repeated sampling. Users can easily port their Prismatic VLM models to SGLang using our provided template. 

We conducted experiments on a single H100 to measure the latency and throughput of OpenVLA inference across varying batch sizes. As shown in Table~\ref{tab:latency_throughput_comparison}, our optimized implementation significantly outperforms the naive version~\cite{kim2024openvla}. For instance, at a batch size of 32, our serving engine reduces latency by 74\% and increases throughput by over 120x. Even at smaller batch sizes (e.g. 4), our VLA serving engine achieves a 54\% reduction in latency. 

Our action verifier also benefits significantly from batch inference, achieving a throughput of 46 actions/s at a batch size of 16. It is notably faster than OpenVLA, as it only requires computation during the prefill stage, which can fully leverage GPU parallelism. In contrast, OpenVLA involves both the prefill and decode stages—where action tokens must be generated autoregressively—resulting in lower throughput.

\section{Trade-off between Action Error and Computational Overhead} \label{H}
\begin{figure*}[h!]
    \centering
    \includegraphics[width=0.7\linewidth]{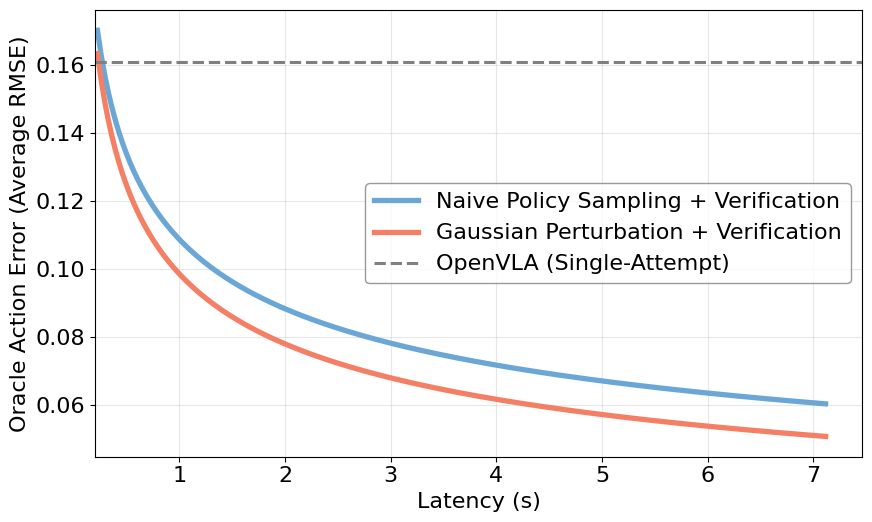}
    \caption{Action error (average RMSE) as a function of computational overhead for policy sampling and Gaussian perturbation. Gaussian perturbation consistently achieves lower action error under equivalent computational budgets}
    \label{fig:latency_comparison}
\end{figure*}

In this ablation study, we examine the trade-off between action error and computational overhead. In Figure~\ref{fig:latency_comparison}, we reproduce the scaling curve from Section~\ref{sec:scaling}, but plotting action error against latency. Under equivalent computation budgets, Gaussian perturbation achieves significantly lower oracle action error compared to policy sampling. We observe that the gap between the two methods widens as compute budgets increase. This growing performance gap reflects the fact that Gaussian perturbation scales only with the verifier, while policy sampling incurs additional overhead from both the policy and verifier. These results highlight Gaussian perturbation as a more practical choice for deployment.

\section{Notation}
\begin{table}[ht]
\centering
\begin{tabular}{ll}
\toprule
\textbf{Symbol} & \textbf{Description} \\
\midrule
$\mathcal{S}$ & State space \\
$\mathcal{A}$ & Action space \\
$\tau$ & Trajectory \\
$I$ & Task instruction \\
$\pi_\theta$ & Generalist robot policy parameterized by $\theta$ \\
$a_t$ & Action at timestep $t$: $[\Delta x_t, \Delta y_t, \Delta z_t, \Delta u_t, \Delta v_t, \Delta w_t, g_t]$ \\
$g_t$ & Binary gripper state (open or close) \\
$P(s^\prime | s, a)$ & Transition dynamics \\
$R_\phi(s, a, I)$ & Reward model parameterized by $\phi$ \\
$\mathcal{D}$ & Demonstration dataset \\
$\mathcal{D}_{\text{buf}}$ & Auxiliary dataset of state-action-instruction tuples \\
$\mathcal{D}_{\text{comp}}$ & Synthetic action preference dataset \\
$\sigma$ & Sigmoid function \\
$N, K$ & Number of candidate and clustered actions (training) \\

$a_t^{i}$ & $i$-th action sampled from policy (training) \\
$\hat{N}, \hat{K}$ & Number of sampled and augmented actions (test-time) \\
$\hat{a}_t^{i}$ & $i$-th action sampled from policy (test-time) \\
$\tilde{a}_t^{i}$ & $i$-th action sampled from proposal distribution \\
$\mu_t, \Sigma_t$ & Mean and covariance of a Gaussian distribution \\
$(a^W_t, a^L_t)$ & Preferred and less preferred action pair \\
$\Delta^*_t$ & Ground-truth RMSE difference \\
$\hat{\Delta}_t$ & Predicted RMSE difference \\
$\alpha$ & Margin weight in loss function \\
\bottomrule
\end{tabular}
\label{tab:notation}
\end{table}

\end{document}